\def\year{2019}\relax
\documentclass[letterpaper]{article} 
\usepackage{aaai19}  
\usepackage{times}  
\usepackage{helvet}  
\usepackage{courier}  
\usepackage{url}  
\usepackage{graphicx}  
\PassOptionsToPackage{numbers, compress}{natbib}

\usepackage{xcolor}
\usepackage{soul}
\usepackage[small]{caption}
\usepackage[export]{adjustbox}
\usepackage{amssymb}
\usepackage{amsthm}
\usepackage{amsmath}
\usepackage{comment}
\usepackage{algorithm}
\usepackage{algorithmic}
\usepackage{amsmath}
\usepackage{moresize}
\usepackage{wrapfig}
\usepackage{subcaption}

\usepackage[utf8]{inputenc} 
\usepackage[T1]{fontenc}    
\usepackage{url}            
\usepackage{booktabs}       
\usepackage{amsfonts}       
\usepackage{nicefrac}       
\usepackage{microtype}      
\def\i#1{\hbox{\it #1\/}}
\def\beq{\begin{equation}}
\def\eeq#1{\label{#1}\end{equation}}
\def\ba{\begin{array}}
\def\ea{\end{array}}

\def\t{\textbf{t}}
\def\f{\textbf{f}}

\def\iif{\hbox{\bf if}}

\def\causes{\hbox{\bf causes}}
\def\inertial{\hbox{\bf inertial}}
\def\default{\hbox{\bf default}}

\def\nonex{\hbox{\bf nonexecutable}}

\newcounter{thm_counter}

\newtheorem{theorem}[thm_counter]{Theorem}

\usepackage[inline,ignoremode]{trackchanges}
\addeditor{lb}

\addeditor{fk}

\addeditor{dm}

\title{SDRL: Interpretable and Data-efficient Deep Reinforcement Learning\\Leveraging Symbolic Planning}

\author{
Daoming Lyu$^1$, 
Fangkai Yang$^2$, 
Bo Liu$^1$,
Steven Gustafson$^3$
\\ 
$^1$ Auburn University, Auburn, AL, USA\\
$^2$ NVIDIA Corporation, Redmond, WA, USA\\
$^3$ Maana Inc., Bellevue, WA, USA  \\
daoming.lyu@auburn.edu,
fangkaiy@nvidia.com,
boliu@auburn.edu,
sgustafson@maana.io
}

\begin{document}

\maketitle

\begin{abstract}
Deep reinforcement learning (DRL) has gained great success by learning directly from high-dimensional sensory inputs, yet is notorious for the lack of interpretability. 
Interpretability of the subtasks is critical in hierarchical decision-making as it increases the transparency of black-box-style DRL approach and helps the RL practitioners to understand the high-level behavior of the system better.
In this paper, we introduce symbolic planning into DRL and propose a framework of Symbolic Deep Reinforcement Learning (SDRL) 
that can handle both high-dimensional sensory inputs and symbolic planning. The task-level interpretability is enabled by relating symbolic actions to options.
This framework features a {\em planner -- controller -- meta-controller} architecture, which takes charge of subtask scheduling, data-driven subtask learning, and subtask evaluation, respectively.  The three components cross-fertilize each other and eventually converge to an optimal symbolic plan along with the learned subtasks, bringing together the advantages of long-term planning capability with symbolic knowledge and end-to-end reinforcement learning directly from a high-dimensional sensory input. 
Experimental results
validate the interpretability of subtasks, along with improved data efficiency compared with state-of-the-art approaches. 
\end{abstract}

\section{Introduction}
%
Deep reinforcement learning (DRL) algorithms have achieved tremendous successes on sequential decision-making problems involving high-dimensional sensory inputs such as Atari games~\cite{dqn:nature:2015}. The input states of Atari games are usually raw pixel images, and a deep neural network is used to approximate Q-values, i.e., ``Deep Q-Network'' (DQN). This approach can learn fine granular policies that surpass human experts but is criticized for the lack of data efficiency and interpretability. DRL algorithms usually require several millions of samples but still cannot learn long-horizon sequential actions for problems with sparse feedback and delayed rewards, such as Montezuma's Revenge \cite{dqn:nature:2015}. The learning behavior based on the black-box neural network is nontransparent and hard to explain and understand. The goal of interpretability is to describe the internals of a system in a way that is understandable to humans. In real applications of decision-making, it is instrumental to make the system behavior interpretable to gain the trust from the user and provide insights for their decision-making process \cite{gilpin2018explaining}. Studying on these two topics has received increasing attention from the machine learning community.
Recent study in Neuroscience suggests human plays video games by learning and planning from a high-level object-based representation of a deterministic transition model of the underlying problem \cite{tsividis2017human}. This observation sheds lights on improving data efficiency by introducing a hierarchy over flat MDP problem using subtasks, i.e., a temporal abstraction over a course of primitive actions, and utilizing object representation for learning \cite{kulkarni2016deep} or strategic planning \cite{keramati2018strategic}. Although these approaches achieved some success on data efficiency, none of them reaches the same level of interpretability as much as object-based subtask planning and learning that human performs.  

In this paper, we argue that performing reasoning and planning on explicitly represented knowledge is an effective way to achieve the task-level interpretability and propose to use {\em Symbolic Planning} (SP) \cite{cim08} as a descriptive and intuitive high-level technique to improve the data-efficiency and interpretability of DRL. Symbolic planning has been used to build mobile robots that co-inhabit with human, perform tasks for human and communicate with human for task-relevant information \cite{hanheide2015robot,chen2016planning,khandelwal2017bwibots}, all requiring high-level interpretability of their behavior. Different from RL approaches, a planning agent carries prior symbolic knowledge of objects, properties and how they are changed by executing actions in the dynamic system, represented in a formal, logic-based language such as PDDL \cite{mcdermott1998pddl} or an action language \cite{gel98} that relates to logic programming under answer set semantics (answer set programming) \cite{lif08}. The agent utilizes a symbolic planner, such as a PDDL planner \textsc{FastDownward}~\cite{helmert2006fast} or an answer set solver \textsc{Clingo}~\cite{gekasc12c} to generate a sequence of actions based on its symbolic knowledge, executes the actions to achieve its goal, perform execution monitor and replan to handle execution failure and domain uncertainty. Compared with RL approaches, an SP agent does not require a large number of trial-and-error to behave reasonably well, and the interpretability of the agent's behavior is well guaranteed by white-box algorithms of planning and reasoning with predefined and human-readable symbolic knowledge. 
Finally, recent work on integrating symbolic planning with RL \cite{yang:peorl:2018,lu2018robot} provides SP+RL frameworks where symbolic plans based on prior knowledge can guide RL for meaningful exploration, leading to improvement on data efficiency for decision making on problems in tabular representations.

This paper advances the state of the art SP+RL approach by integrating symbolic planning with a hierarchical approach of DRL. We made two assumptions. First, it is relatively easy for human expert to represent knowledge about high-level abstract, albeit inaccurate, dynamics of the problem domain. Second, due to the recent progress on computer vision, it is relatively easy to build perception modules to recognize objects and their properties. Based on the assumptions, we propose the {\em intrinsic goal}, a measurement of plan quality based on an internal utility function, to enable reward-driven planning. Afterwards, we propose a {\em Symbolic Deep Reinforcement Learning} (SDRL) framework that features a {\em planner--controller--meta-controller} architecture, i.e.,
\begin{enumerate}
\item A planner uses prior symbolic knowledge to perform long-term planning by a sequence of symbolic actions (subtasks) that achieve its intrinsic goal; 
\item A controller uses DRL algorithms to learn the sub-policy for each subtask  based on {\em intrinsic rewards};
\item     A meta-controller learns on {\em extrinsic rewards} by measuring the training performance of controllers and propose new intrinsic goals to the planner.
\end{enumerate}
In this process, planner, controllers, and meta-controller cross-fertilize each other and eventually converge to an optimal symbolic plan along with the learned subtasks.
To the best of our knowledge, this is the first work that integrates symbolic planning with DRL that gains interpretability and data-efficiency for decision-making in complex domains. While our framework is generic enough so that various planning and DRL techniques can be used, we instantiate our framework using action language ${\cal BC}$ \cite{lee13} for planning and R-learning \cite{rlearning:schwartz} for meta-controller learning. 
We prove that the symbolic plan converges to optimal conditioned on the convergence of meta-controller. The framework is evaluated on Taxi domain \cite{barto-sm:hrl} and Montezuma's Revenge, demonstrating improved interpretability through explicitly encoding planning knowledge and learning into human-readable subtasks, and also improved data-efficiency through automatic selecting and learning control policies of modular subtasks.

\section{Preliminaries}\label{sec:prelim}
\textbf{Planning with Action Language ${\cal BC}$.} An {\em action description}~$D$ in the language ${\cal BC}$ \cite{lee13} includes two kinds of symbols on signature $\sigma$, {\em fluent constants} that represent the properties of the world, and {\em action constants}. A \textit{fluent atom} is an expression of the form $f=v$, where $f$ is a fluent constant and $v$ is an element of its domain. For boolean domain, denote $f=\t$ as $f$ and~$f=\f$ as $\sim\!\!\! f$. An action description is a finite set of {\em causal laws} that describe how fluent atoms are related with each other in a single time step, or how their values are changed from one step to another, possibly by executing actions. For instance,
$
(A~\iif~A_1,\ldots,A_m)
$
is a {\em static law} that states at a time step, if $A_1,\ldots, A_m$ holds then $A$ is true. Another static law
$
(\default~f=v)
$
states that by default, the value of $f$ equals~$v$ at any time step. 
$
(a~\causes~A_0~\iif~A_1,\ldots, A_m)
$
is a {\em dynamic law}, stating that at any time step, if $A_1,\ldots, A_m$ holds, by executing action $a$,~$A_0$ holds in the next step. 
$
(\nonex~a~\iif~A_1,\ldots,A_m)
$
states that at any step, if $A_1,\ldots, A_m$ holds, action $a$ is not executable. Finally, the dynamic law
$
(\inertial~f)
$
states that by default, the value of fluent $f$ does not change with time. 

An action description captures a dynamic transition system. A {\em symbolic state} $s$ is a complete set of fluent atoms, and a transition is a tuple $\langle s,a,s' \rangle$ where $s, s'$ are states and~$a$ is an action. 
Let $I$ and $G$ be states. The triple $(I,G,D)$ is called a planning problem. $(I,G,D)$ has a plan of length $l-1$ iff there exists a transition path of length $l$ such that $I=s_1$ and $G=s_l$. Throughout the paper, we use $\Pi$ to denote both the plan and the transition path by following the plan. Due to the semantic definition above, automated planning with an action description in ${\cal BC}$ can be achieved by an answer set solver such as \textsc{Clingo}, and the output answer sets encode the transition paths that solve the planning problem. 

\noindent\textbf{Reinforcement Learning.} A Markov Decision Process (MDP) is defined as the tuple $({\mathcal{S},\mathcal{A},P_{ss'}^{a},r,\gamma})$, where $\mathcal{S}$ and $\mathcal{A}$ are the sets of symbols denoting states and actions, the transition kernel $P_{ss'}^{a}$ specifies the probability of transition from state $s\in\mathcal{S}$ to state $s'\in\mathcal{S}$ by taking action $a\in\mathcal{A}$, $r(s,a):\mathcal{S}\times\mathcal{A}\mapsto\mathbb{R}$ is a reward function bounded by $r_{\max}$, and $0\leq\gamma<1$ is a discount factor. A solution to an MDP is a policy $\pi:\mathcal{S}\mapsto \mathcal{A}$ that maps a state to an action. RL learns a near-optimal policy  by executing actions and observing the state transitions and rewards, and can be applied when the underlying MDP is not known.

To evaluate a policy $\pi$, there are two types of performance measures: the expected discounted sum of reward for infinite-horizon problems 
and the expected un-discounted sum of rewards for finite horizon problems. In this paper we adopt the latter metric defined as $J^\pi_{\rm avg}(s) = \mathbb{E}[\sum\limits_{t = 0}^T {{r_t}}|s_0=s ]$, where $T$ denotes the horizon length. We define the \textit{gain reward} ${\rho ^\pi }(s)$ reaped by policy $\pi$ from $s$ as
$
{\small
{\rho ^\pi }(s) = \mathop {\lim }\limits_{T \to \infty } \frac{{J^\pi_{{\rm{avg}}}(s)}}{T} = \mathop {\lim }\limits_{T \to \infty } \frac{1}{T}\mathbb{E}[\sum\limits_{t = 0}^T {{r_t}} ]
} .
$
R-learning \cite{sm:mlj96} 
 is a model-free value iteration algorithm that can be used to find the optimal policy for average reward  criteria. At the $t$-th iteration $(s_t,a_t,r_t, s_{t+1})$, update:
\begin{align}
 {R_{t + 1}}({s_t},{a_t})\!\! & \xleftarrow{\alpha_t} {r_t} - {\rho _t}(s_t)  
 + \mathop {\max }\limits_a {R_t}({s_{t + 1}},a), \nonumber \\
 \rho_{t+1}(s_t)\!\!& \xleftarrow{\beta_t} r_t + \mathop {\max }\limits_a {R_t}({s_{t + 1}},a) - \mathop {\max }\limits_a {R_t}({s_t},a) \nonumber
\end{align}
where $\alpha_t, \beta_t$ are the learning rates, and $a_{t+1} \xleftarrow{\alpha} b$ denotes the update law as ${a_{t + 1}} = (1-\alpha){a_{t}} + \alpha b$.

\noindent\textbf{Options.} Compared with regular RL, hierarchical reinforcement learning (HRL) \cite{barto-sm:hrl} specifies on real-time-efficient decision-making problems over a series of tasks.
 An MDP can be considered as a flat decision-making system where the decision is made at each time step. On the contrary, humans make decisions by incorporating temporal abstractions.
An {\em option} is temporally extended course of action consisting of three components: a policy $\pi: {\mathcal{S}} \times {\mathcal{ A }} \mapsto {{ [0, 1]}}$, a termination condition~$\beta: {\mathcal{S}} \mapsto {{ [0, 1]}}$, and an initiation set ${\mathcal{I}} \subseteq {\mathcal{S}}$. An option~$(I,\pi ,\beta)$ is available in state $s_t$ iff ${s_t} \in I$. After the option is taken, a course of actions is selected according to~$\pi$ until the option is terminated stochastically according to the termination condition $\beta$.
With the introduction of options, the decision-making has a hierarchical structure with two levels, where the upper level is the option level (also termed as task level) and the lower level is the (primitive) action level. Markovian property exists among different options at the option level. 

\section{Related Work}\label{sec:related}
\noindent{\bf Interpretability.} Studying on interpretability concerns on describing the internals of a
system in a way that is understandable to humans \cite{doshi2017towards,gilpin2018explaining}. Interpretability of deep learning involves studying on explaining deep network processing \cite{ribeiro2016should}. 
For interpretive DRL, program induction approach is used \cite{verma2018programmatically} to enable policy interpretability. Our work is the first to use symbolic knowledge to enable task-level interpretability for DRL and it is interpretable by construction.

\noindent{\bf Hierarchical Deep Reinforcement Learning.}
Hierarchical RL approach such as the options framework~\cite{hrl:option:sutton1999} formulates the problem using a two-level hierarchy as aforementioned and is one way to solve the challenge of learning long horizon action sequences with sparse rewards. It often assumes that a set of useful options are predefined. 
\cite{machado2017laplacian,machado2017eigenoption} focus on discovering eigen-based options and also attempt to solve the problem of learning policies over long time horizons. However, it is difficult to interpret the options in their approaches. The closest hierarchical RL work to ours are \cite{kulkarni2016deep,le2018hierarchical}, utilizing a meta-controller to learn to sequence subtasks defined on objects.
In addition, \cite{keramati2018strategic} focuses on performing strategic planning and learning in Object-oriented MDP, a deterministic abstraction over the original problem. 
By contrast, we leverage symbolic action languages to explicitly represent objects, properties, and the high-level transition dynamics. We use an out-of-box symbolic planner to generate and improve plans, with symbolic transitions automatically mapped to subtasks, leading to a more interpretable and expressive representation. 

\noindent{\bf Integrating Symbolic Planning with Reinforcement Learning.}
SP-RL integration has been studied for a long time \cite{parr1998reinforcement,Ryan02usingabstract,hogg2010learning,leonetti2016synthesis,yang:peorl:2018,lu2018robot}, most of which are based on tabular representation of the domain. 
Our work inherits the interpretability of symbolic planning with symbolic knowledge and further generalizes previous PEORL framework \cite{yang:peorl:2018} with intrinsic goals and the integration with DRL. In particular, the meta-controller is introduced to bridge the gap of \textit{planning over symbolic states and DRL over pixel images}, by learning at the task level using extrinsic reward derived from training performance of DRL. Meta-controller learning enables the planner to perform automatic selection on subtasks and improve the plan by sequencing learnable subtasks. 

\noindent{\bf Computational Models of Intrinsic Motivation.}
Recent studies \cite{kulkarni2016deep} showed that characterizing {\em intrinsic motivation} is important to address learning goal-directed behavior facing sparse feedback and delayed rewards. In psychology, intrinsic motivation is defined as accomplishing an activity for its inherent satisfaction rather than for some separable consequences, driven by an internal utility function \cite{oudeyer2009intrinsic}. Intrinsically-motivated RL \cite{intrinsic:barto:2005} uses the framework of options. In comparison, our work provides a computational model where symbolic planning uses its internal utility function to measure its plan quality so that this plan quality can motivate the agent to improve the plan by accumulating larger rewards.

\section{SDRL Framework}
\label{sec:sdqnim}
We model the underlying sequential decision-making problem as an MDP tuple $(\widetilde{\mathcal{S}},\widetilde{\mathcal{A}},\widetilde{P^a_{ss'}},r,\gamma)$ where $\widetilde{\mathcal{S}}$ consists of states of high-dimensional sensory inputs such as pixel images, $\widetilde{\mathcal{A}}$ is the set of primitive actions, $\widetilde{P^a_{ss'}}$ is the transition matrix, $r$ is the reward function, and $\gamma$ is a discounting factor. 
In the following paper, $\widetilde{\mathcal{S}}, \widetilde{\mathcal{A}}$ are used to denote the MDP state space and action space, while $\mathcal{S}, \mathcal{A}$ represent the symbolic state space and action space.
Our goal is to learn both a sequence of subtasks and the corresponding sub-policies, so that executing the sub-policy for each subtask one by one can achieve maximal cumulative reward. 

To solve this problem, we assume a symbolic structure, i.e., a set of causal rules that captures objects, fluents and how their values are changed by executing subtasks, is given by human experts. While a pre-defined symbolic representation requires some work from human experts, it has been observed that majority of discrete dynamic domains share surprising similarities and can be formulated based on a set of general-purpose action modules \cite{erdo06,inclezan2016modular}, and symbolic representation is elaboration-tolerant: adding new information usually doesn't require significant change of existing knowledge. Consequently, the symbolic formulation for one problem can be easily applied to another, by instantiating a different set of objects or adding a few more rules.  For instance, Taxi domain, a benchmark problem of HRL, concerns the movement of a taxi in a gridworld, carrying passengers and dropping off passenger at destination, while Montezuma's Revenge, a seemingly drastically different Atari game, concerns the movement of an Avatar among a set of locations (ladders, platforms, doors, ropes, etc), picking up a key and using the key to open a door. Both domains involve the formulation of spatial movement, co-location of objects and utility of objects. Furthermore, our framework is intended to start with a coarse granular, high-level abstract domain formulation, so that decision-making can be robust and flexible facing domain change and uncertainty. Consequently, the laborious effort of crafting an accurate symbolic model is neither necessary nor useful.

\begin{figure}
\centering
\includegraphics[height=3cm,width=7cm]{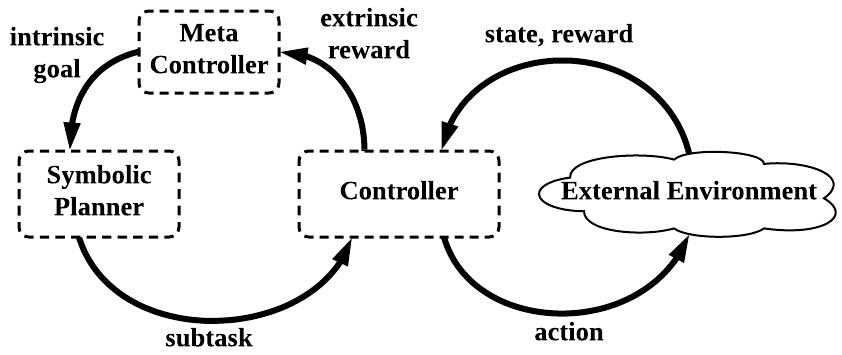}\\
\caption{Architecture illustration}
\label{fig:arch}
\end{figure}

With a symbolic representation given by the human expert, the SDRL architecture is shown in Fig.~\ref{fig:arch}. A symbolic planner generates high-level plans, i.e., a sequence of subtasks, to meet its {\em intrinsic goal}. An intrinsic goal is a measurement on plan quality, which approximates how much cumulative reward the plan may achieve. 
We assume a pre-trained function can associate each sensory input with a symbolic state, i.e., performing symbol grounding, 
so that subtasks on the MDP space can be induced based on symbolic states and the pre-trained function. We extend the reward structure of core MDP by introducing {\em intrinsic reward} and {\em extrinsic reward} to facilitate two levels of learning tasks. The sub-policies for the action level are learned using DRL algorithms based on intrinsic reward, with pseudo-rewards to encourage the agent to learn skills to achieve each subtask. As DRL continues, a metric is used to evaluate the competence of learned sub-policies, such as the success ratio over a number of episodes, from which extrinsic rewards is derived. When the sub-policy is learned and reliably achieves the subtask, the extrinsic reward is equivalent to the environmental reward. Using extrinsic rewards, meta-controller performs R-learning that reflects the long-term average reward and gains the reward of selecting each subtask. The learned values are returned to the symbolic planner, and used to measure plan quality and propose new intrinsic goals for the planner to improve the plan, by either exploring new subtasks or by sequencing learned subtasks that supposedly can achieve higher rewards in the next iteration. We define the process formally as follows.

\subsection{Symbolic Representation}\label{sec:sr}

A tuple $(I, G, D)$ is a symbolic planning problem, where $D$ is an {\em  action description} defined on signature $\sigma$, $I$ is initial state and $G$ is the {\em intrinsic goal}. 

Similar to PEORL, we use action language ${\cal BC}$ to demonstrate the essence of the action description, but similar formulation can be represented easily using other planning languages such as PDDL. In addition to usual causal laws that describe the preconditions and effects of actions (dynamic laws) and static relationships between fluents (static law), $D$ consists of causal laws that formulate gain rewards of executing actions and its effect on cumulative plan quality:
\begin{itemize}
\item For any symbolic state that contains atoms $\{A_1,\ldots, A_n\}$, $D$ contains static laws of the form:  
$
 s~\iif~A_1,\ldots, A_n,~\hbox{for~state}~s\in\mathcal{S}.
$
\item We introduce new fluent symbols of the form $\rho(s,a)$ to denote the gain reward at state $s$ following action $a$. $D$ contains a static law stating by default, the gain reward is initialized optimistically, denoted as~$\i{INF}$, to promote exploration:
$
\default~\rho(s,a)=\i{INF},~\hbox{for}~s\in\mathcal{S},a\in\sigma_A(D).
$
\item We use fluent symbol $\i{quality}$ to denote the cumulative average-adjusted reward of a plan, termed as {\em plan quality}. $D$ contains dynamic laws of the form:
$
a~\causes~\i{quality}=C+Z~\iif~s,\rho(s,a)=Z,\i{quality}=C.\!
$
\item $D$ contains a set $P$ of facts of the form $\rho(s,a)=z$.
\end{itemize}
$I$ is the initial symbolic planning state, and $G$ is an {\em intrinsic goal} which is a linear constraint of the form 
\beq\i{quality} > \i{quality}(\Pi),
\eeq{goal}
for a symbolic plan $\Pi$ measured by the internal utility function $\i{quality}$ defined as
\beq
\i{quality}(\Pi) = \sum_{\langle s_{i},a_{i},s_{i+1}\rangle\in\Pi} \rho(s_{i},a_{i}).
\eeq{quality}
The definition of intrinsically motivated goal~(\ref{goal}) is different from standard PEORL, in which a goal consists of the linear constraint of form (\ref{goal}) plus a set of logical constraints specifying the goal condition, given by the human designer towards a particular task. Intrinsic goal in SDRL drops the logical constraint part and enables ``model-based exploration by planning'', which is more suitable for RL problems where the agent's behavior is driven by reward.

\subsection{From Symbolic Transitions to Options}
Given the set $\mathcal{S}$ of symbolic states, i.e., a complete set of fluent atoms,  we assume there is a pre-trained oracle capable of answering whether the symbolic properties specified as fluent atoms of the form $f=v$ in $s$ are true in the high-dimensional sensory input $\tilde{s}$, and define the mapping $\mathbb{F}$ as $\mathbb{F}:\mathcal{S}\times\widetilde{\mathcal{S}}\mapsto\{\t,\f\}$. In the case of Atari games, such pre-trained function can be a perception module that performs object recognition and performs symbol grounding based on the predefined semantics of symbols. For instance, the perception module can answer if the avatar picked the key by checking if the bounding box of the avatar overlaps with the bounding box of the key. Due to the recent progress of computer vision, we assume such a perception module is generally available.

Given $\mathbb{F}$ and a pair of symbolic states $s,s'\in\mathcal{S}$, we can induce a semi-Markov option, as a triple $(I,\pi,\beta)$ where the initiation set $I=\{\tilde{s}\in\widetilde{\mathcal{S}}:\mathbb{F}(s,\tilde{s})=\t\}$, $\pi:\widetilde{\mathcal{S}}\mapsto\widetilde{\mathcal{A}}$ is the intra-option policy, and $\beta$ is the termination condition such that 
$\beta(\tilde{s'})=\left\lbrace
\ba{ll}
1 & \mathbb{F}(s',\tilde{s'})=\t, \hbox{for}~\tilde{s'}\in\widetilde{\mathcal{S}}\\
0 &\hbox{otherwise}
\ea\right.
$.
The formulation above maps symbolic transition to a similar structure of options.

\subsection{Rewards}
To facilitate learning at the action level and the task level, we define intrinsic reward at the action level as

\beq
r_i(\tilde{s'})=\left\lbrace
\ba{ll}
\phi & \beta(\tilde{s'})=1\\
r & \hbox{otherwise}
\ea\right.
\eeq{intrinsic}
where $\phi$ is a large number encouraging achieving subtasks and $r$ is the reward from the environment at state $\tilde{s'}$. If reward is sparse, (\ref{intrinsic}) is usually a simple binary form.
We further define extrinsic reward for selecting subtask $g$ at symbolic state $s$ as
$
r_e(s,g) = f(\epsilon)
$
where $f$ is a function about $\epsilon$, a criterion that measures the competence of the learned sub-policy for each subtask. 
We define $\epsilon$ as the success ratio, which is the average rate of successfully completing the subtask over the previous 100 episodes.
$f$ can be defined as
\beq
f(\epsilon)=\left\lbrace
\ba{ll}
-\psi & \epsilon<0.9\\
r(s,g) & \epsilon\ge 0.9
\ea\right.
\eeq{extrinsic}
where $\psi$ is a large number to punish selecting unlearnable subtasks, $r(s,g)$ is the discounted cumulative reward obtained from the environment by following the subtask $g$, and $0.9$ is the threshold. 
Unlearnable sub-tasks here refers to the sub-tasks that are too difficult to learn by the controller on the condition that the success ratio of achieving a sub-task cannot keep above the threshold value of 0.9 after the training by episodes.
Intuitively, the definition of extrinsic rewards means if the sub-policy can reliably achieve the subgoal, then the extrinsic reward at $s'$ reflects true cumulative environmental reward of following the subtask; otherwise, the extrinsic reward at $s'$ is negative, indicating that the sub-policy performs badly and is probably not learnable.

A plan $\Pi$ of $(I,G,D)$ is considered to be {\em optimal} iff
$\sum_{\langle s,a,s'\rangle} r_e(s, g)$ is maximal among all plans.


\subsection{Planning and Learning}
The planning and learning process is shown in Algorithm~\ref{algexec}. At any episode $t$, symbolic planner uses a logical representation $D$, an initial state $I$, and an intrinsic goal $G$ to generate a symbolic plan $\Pi_t$ (Line 4).The symbolic transitions of $\Pi_t$ correspond to subtasks (Line 10) and sent to a controller to learn the sub-policy for each subtask over a predefined number of steps (Line 11). The controller performs deep Q-learning with intrinsic rewards $r_i$ using experience replay (Lines 12--15). The controller estimates the $Q$ value $Q(\tilde{s},\tilde{a};g)\approx Q(\tilde{s},\tilde{a};\theta,g)$, where $\theta$ is the parameter of the non-linear function approximator. The experience of executing actions in the environment $(\tilde{s}_t,\tilde{a}_t,r_e(\tilde{s}_{t+1},g),\tilde{s}_{t+1})$ is stored in memory $\mathcal{D}_g$, and the loss function is defined as
\beq
\ba{l}
L(\theta;g) = \hbox{E}_{(\tilde{s},\tilde{a},g,r_i,\tilde{s}')\sim D_g}\\
~~~~[r_e + \gamma\max_{\tilde{a'}} Q(\tilde{s},\tilde{a'};\theta_{i-1},g)- Q(\tilde{s},\tilde{a};\theta_i,g)]^2
\ea
\eeq{dqn}
where $i$ denotes the iteration number. After maximal steps are reached, the success ratio of controller's sub-policy or the true environmental rewards are used to derive extrinsic rewards (Line 17). Meta-controller performs R-learning (Line 18) based on extrinsic rewards for the symbolic transitions $\langle s_t,a_t,s_{t+1}\rangle$ that corresponds to the subtask $g_t$:
\beq
\ba{rl}
R_{t+1}(s_{t},g_{t})\!\!\!&\!\!\xleftarrow{\alpha} r_e - \rho_t^{g_{t}}(s_{t})+\max_{g} R(s_t,g)\\
\rho_{t+1}^{g_{t}}(s_{t})\!\!\!&\!\!\xleftarrow{\beta}r_e+\max_{g} R_t(s_{t+1},g) - \max_{g} R_t(s_{t},g)
\ea
\eeq{riter1}
After R-learning is performed, the {\em quality} of the symbolic plan $\Pi_t$ is measured by (\ref{quality}) (Line 20). The plan quality $\i{quality}(\Pi_t)$ is used to update intrinsic goal (Line 21) and learned~$\rho$ values are passed back into symbolic formulation for a new plan to be generated (Line 22). The loop continues until the symbolic plan $\Pi^{*}$ cannot be further improved. 
\begin{algorithm}[htb!]
{\small
  \caption{SDRL Planning and Learning Loop}
  \label{algexec}
  \begin{algorithmic}[1]
    \REQUIRE $(I,G,D,\mathbb{F})$ where $G=(\i{quality}>0)$, and an exploration probability $\epsilon$
    \STATE Initialization: $P_0\Leftarrow \emptyset$, $\Pi_0\Leftarrow \emptyset$
    \FOR{t = 1 \dots end of episodes} 
      \STATE $\Pi^*\Leftarrow \Pi_{t-1}$
      \STATE take $\epsilon$ probability to solve planning problem and obtain a plan $\Pi_t \Leftarrow\textsc{clingo}.\i{solve}(I,G,D\cup P_{t-1})$
      \IF {$\Pi_t =\emptyset$}
          \RETURN $\Pi^*$
      \ENDIF
      \FOR {symbolic transition $\langle s, a, s'\rangle \in \Pi_t $}
      \STATE obtain current state $\tilde{s}$
      \STATE correspond to subtask $g$ by using $\mathbb{F}$ to obtain initiation set and terminate condition
      \WHILE {$\beta(\tilde{s})\neq 1$ and maximal step is not reached}
      \STATE pick up an action $\tilde{a}$ and obtain transition $(\tilde{s},\tilde{a},\tilde{s'},r_i(\tilde{s'}))$
      \STATE store transition in experience replay buffer $\mathcal{D}_g$
      \STATE estimate $Q(\tilde{s},\tilde{a};\theta,g)$ by minimizing loss function~(\ref{dqn}) when there are sufficient samples in $\mathcal{D}_g$
      \STATE update current state $\tilde{s}\Leftarrow\tilde{s'}$
      \ENDWHILE
      \STATE calculate extrinsic reward $r_e(s,g)$ 
      \STATE update $R(s,g)$ and $\rho^g(s)$ using (\ref{riter1}).
      \ENDFOR
      \STATE calculate quality of $\Pi_t $ by (\ref{quality}).
      \STATE update planning goal $G\Leftarrow (\i{quality}> \i{quality}(\Pi_t))$.
      \STATE update facts $P_t\Leftarrow \{\rho(s,a)=z:\langle s,a,s'\rangle\in\Pi_t, \rho_t^{a}(s)=z\}$
    \ENDFOR
  \end{algorithmic}}
\end{algorithm}

%
The algorithm guarantees symbolic level optimality conditioned on R-learning convergence. See Appendix for proof.
%

%
\begin{theorem}[Termination]
If the meta-controller's R-learning converges, Algorithm~\ref{algexec} terminates iff an optimal symbolic plan exists.
\end{theorem}

%
%
\begin{theorem}[Optimality]
If meta-controller's R-learning converges, when Algorithm~\ref{algexec} terminates, $\Pi^*$ is an optimal symbolic plan.
\end{theorem}


\section{Experiment}\label{sec:exp}
We use Taxi domain \cite{barto-sm:hrl} to demonstrate the behavior of intrinsically motivated planning, and on Montezuma's Revenge~\cite{dqn:nature:2015} for interpretability and data-efficiency. We use 1M to denote $1$ million and 1k to denote $1000$. 

\begin{figure*}[htb!]
\centering
\begin{subfigure}{.28\textwidth}
  \centering
  \includegraphics[width=\linewidth, height=3.6cm]{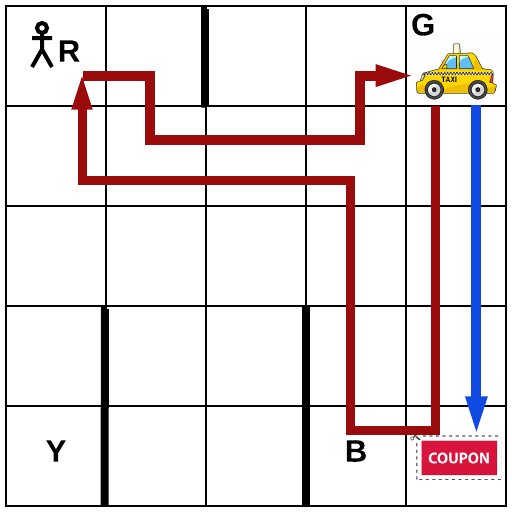}
  \caption{Taxi Domain}
  \label{fig:t2}
\end{subfigure}
\begin{subfigure}{.33\textwidth}
  \centering
  \includegraphics[width=1\linewidth,height=3.8cm]{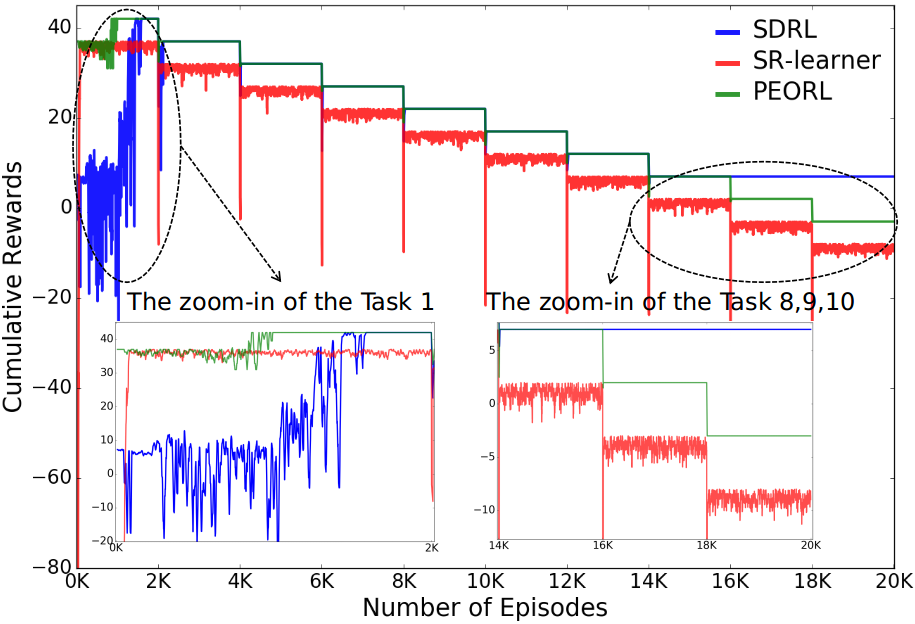}
  \caption{Learning Curve (Taxi Domain)}
  \label{fig:taxi-reward}
\end{subfigure}
\begin{subfigure}{.33\textwidth}
 \includegraphics[width=1\linewidth, height=3.8cm]{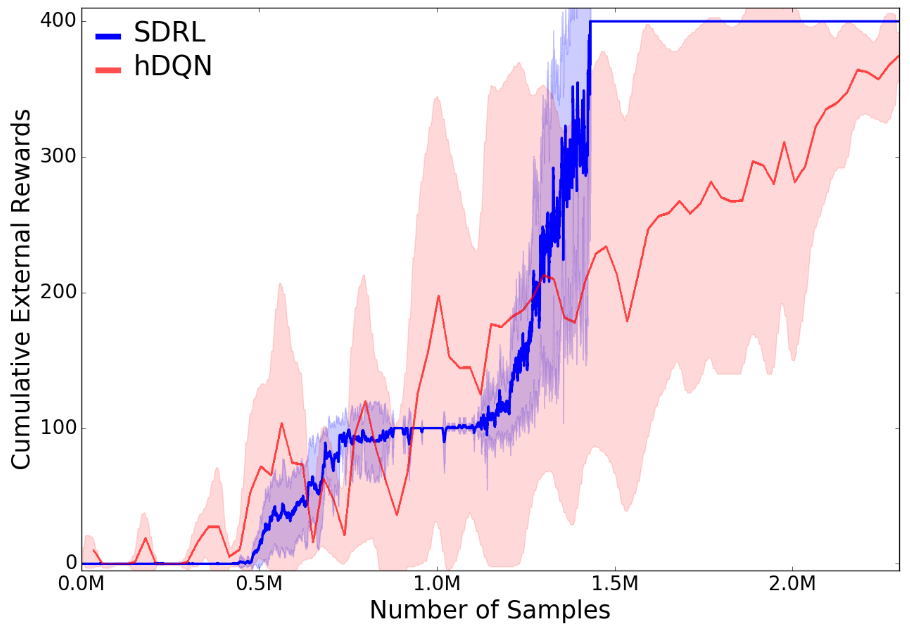}
   \caption{Learning Curve (Montezuma's Revenge)}
   \label{fig:cumulative}
\end{subfigure}
\begin{subfigure}{.33\textwidth}
  \includegraphics[width=1\linewidth, height=3.8cm]{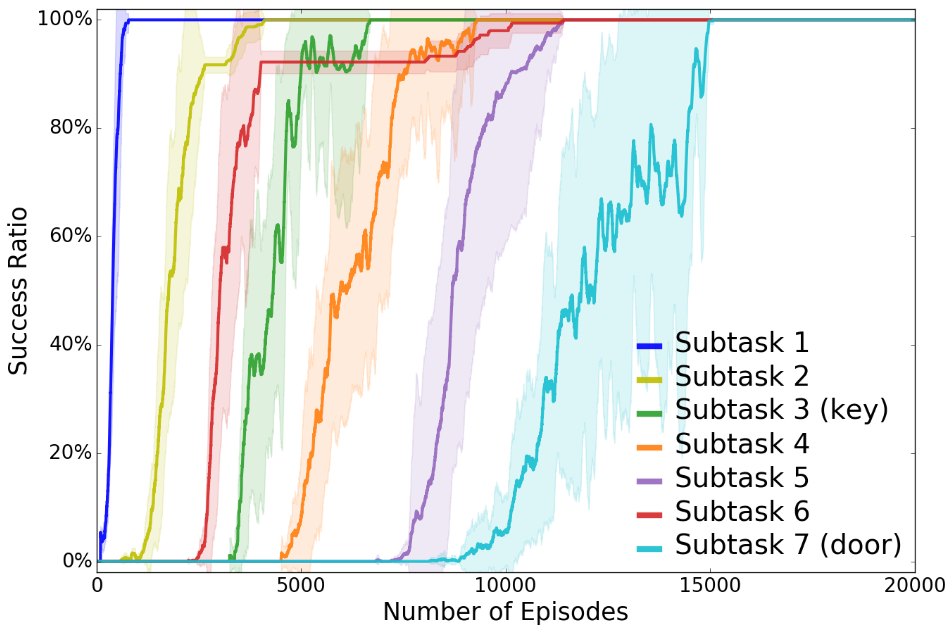}
  \caption{Success Ratio (Montezuma's Revenge)}
  \label{fig:success}
\end{subfigure}
 \begin{subfigure}{.33\textwidth}
\includegraphics[width=1\linewidth, height=3.8cm]{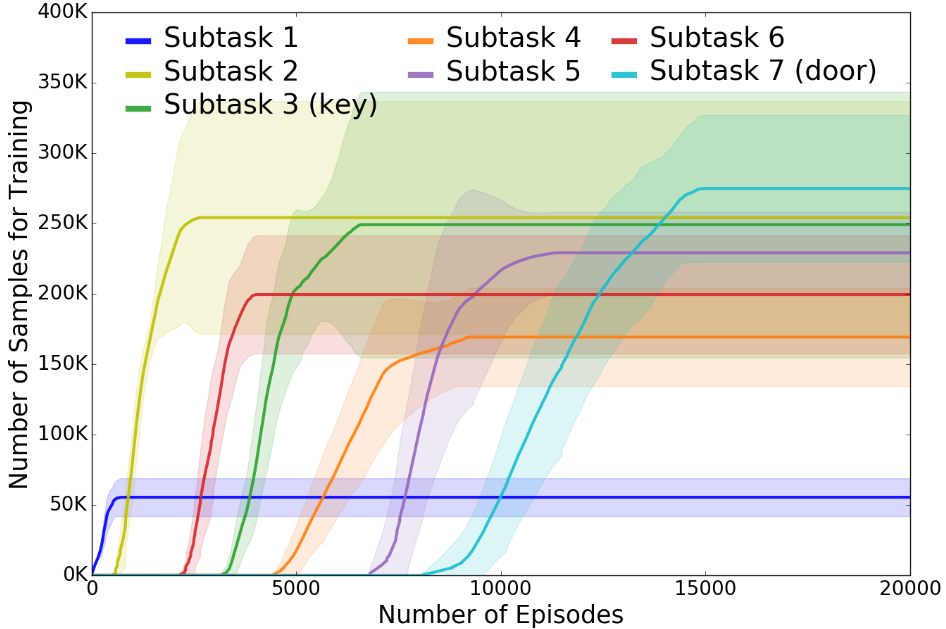}
   \caption{Samples Required (Montezuma's Revenge)}
   \label{fig:sample}
 \end{subfigure}
 \begin{subfigure}{.33\textwidth}
  \includegraphics[width=1\linewidth, height=3.7cm]{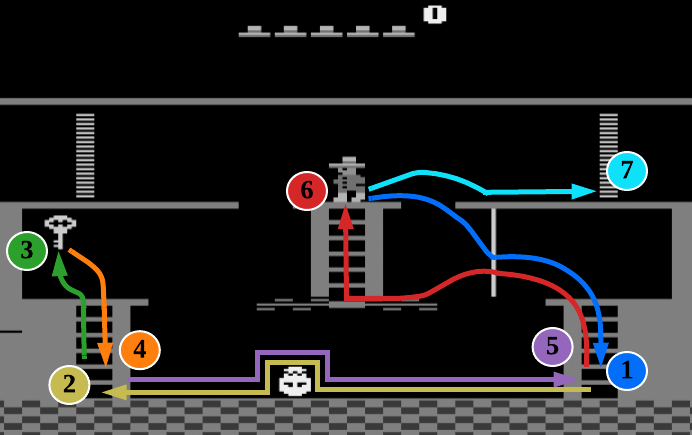}
   \caption{Final Solution (Montezuma's Revenge)}
   \label{fig:final}
\end{subfigure}
\caption{Experimental Results}
\label{fig:taxi}
\end{figure*}


\subsection{Taxi Domain} 
A Taxi starts at any location in a $5\times 5$ grid map (Fig.~\ref{fig:t2}) with a passenger and a destination. 
Every movement has a reward $-1$. Successful drop-off receives reward $50$. Improper pick-up or drop-off receive a reward $-10$. We introduce an extra coupon at $(4,4)$ where the taxi can only collect once, with gaining a reward of $10$. In tabular representation, the intrinsic goal is the only difference between SDRL and standard PEORL, and we will demonstrate how intrinsically motivated goal affects exploration.

\noindent\textbf{Setup.} We consider a sequence of 10 tasks. For each task, the reward of successfully dropping off the passenger declines by 5. For example, the reward of successful dropping off the passenger in Task 1 is $50$, while that of Task 2 would be $45$. Reward change happens after every 2000 episodes, and the taxi's location is always reset to $(0,4)$. Standard PEORL has a fixed final goal, i.e., drop off the passenger at the destination. We compare SDRL with standard PEORL and a linear successor representation (SR) learner \cite{lehnert2017advantages}, a common approach to implement transferable RL for tasks with reward change. 

\noindent\textbf{Experimental Results.} Result is collected and averaged over 10 runs, and learning curves of cumulative reward are shown in Fig.~\ref{fig:taxi-reward}. From Task 1 to Task 7, the optimal policy is to pick up the coupon and then drop off the passenger (the route with dark red color in Fig.~\ref{fig:t2}). Both Standard PEORL and SDRL can learn the optimal policy. As the zoom-in of the Task 1 (first 2k episodes) shown in Fig.~\ref{fig:t2}, SDRL does not converge as fast as SR-learner initially, due to the fact that the exploration of SDRL based on planning with intrinsic goals is not as aggressive as model-free exploration. SDRL does not converge as fast as standard PEORL either because SDRL is not given the explicit planning goal of dropping off the passenger and it needs to explore the states to discover the reward, unlike standard PEORL. SR-learner performs the worst by only dropping off the guest without picking up the coupon. After exploring the state space in Task 1, from Task 2 onwards, SDRL converges as fast as SR-learner and standard PEORL.  However, as the reward of dropping off the passenger keeps declining, from Task 8 onwards, the optimal policy changes to just pick up the coupon. The new optimal policy is successfully learned by SDRL due to intrinsic goal (the route with blue color in Fig.~\ref{fig:t2}), while both standard PEORL and SR learner does not change its policy from previously learned ones. The inadequacy of SR in transferring to an optimal policy with a different goal was also pointed out in \cite{lehnert2017advantages}. Standard PEORL cannot change its plan due to the fixed planning goal, showing that an explicitly given planning goal may unnecessarily restrict exploration of RL, while intrinsic goals in SDRL allows the agent to flexibility changes its goal based on reward structure of the tasks, which is more suitable to solve RL problems.

\subsection{Montezuma's Revenge. }
``Montezuma’s Revenge'' requires the player to navigate the explorer through several rooms while collecting treasures. For the first room, the player has to first pick up the key by climbing down the ladders and moving towards the key in order to pass through doors, resulting in
a long sequence of actions before receiving a reward for collecting the key ($+100$). After that, the player has to move towards the door and open it, which results in another reward ($+300$). Optimal execution requires more than $200$ primitive
actions. Vanilla DQN
frequently achieves a score of $0$ on this domain~\cite{dqn:nature:2015}.

\begin{table}[htb]
{\scriptsize
\begin{center}
\begin{tabular}{ c|c| c }
\hline\hline No.& Layer &  Details  \\
  \hline
1&  Convolutional Layer & 32 filters, kernel size=8, stride=4, activation='relu' \\
2& Convolutional Layer & 64 filters, kernel size=4, stride=2, activation='relu' \\
3& Convolutional Layer & 64 filters, kernel size=3, stride=1, activation='relu' \\
4&  Fully Connected Layer & 512 nodes, activation='relu'\\
5&  Output Layer & activation='linear' \\
\hline\hline
\end{tabular}
\end{center}}
\caption{Neural Network Architecture for Montezuma's Revenge}
 \label{tab:montezuma-nn}
\end{table}

\begin{figure}
{\scriptsize
\begin{verbatim}
% object declaration
location(mp;rd;ls;lll;lrl;key).
% dynamic causal law declaration
move(L) causes loc=L if location(L).
move(L) causes cost=L+Z if rho((at(L1)),move(L))=Z,
           loc=L1,picked(key)=false.
move(L) causes cost=L+Z if rho((at(L1),picked(key)),
           move(L))=Z,loc=L1,picked(key)=true.
inertial loc. inertial quality.
% static causal law declaration
picked(key)=true if loc=key.
nonexecutable move(key) if picked(key).
default rho((at(L1)),move(L))=10.     
default rho((at(L1),picked(key)),move(L))=10.
\end{verbatim}}
\caption{Montezuma's Revenge in ${\cal BC}$}\label{kr}
\end{figure}

\begin{table}[htb!]
{\scriptsize
\begin{center}
\begin{tabular}{ c|c| c| c }
\hline\hline No.& subtask &  policy learned & in optimal plan  \\
  \hline
1&  MP to LRL, no key & \checkmark & \checkmark \\
2&  LRL to LLL, no key & \checkmark & \checkmark\\
3&  LLL to key, no key & \checkmark &\checkmark\\
4&  key to LLL, with key& \checkmark & \checkmark \\
5&  LLL to LRL, with or without key & \checkmark & \checkmark\\
6&  LRL to MP, with or without key & \checkmark & \checkmark \\
7&  MP to RD, with key & \checkmark & \checkmark\\
\hline
8&  LRL to LS, with or without key & \checkmark &  \\
9&  LS to key, with or without key & \checkmark &  \\
10& MP to RD, no key & \checkmark &  \\
\hline
11& LRL to key, with or without key &  &  \\
12 & key to LRL, with key &  &   \\
13 & LRL to RD, with key &  &  \\
\hline\hline
\end{tabular}
\end{center}}
\caption{Subtasks for Montezuma's Revenge}
 \label{tab:montezuma-subgoal}
\end{table}

\noindent{\bf Setup.} Our experiment setup follows the DQN controller architecture~\cite{kulkarni2016deep} with double-Q learning \cite{van2016deep} and prioritized experience replay \cite{schaul2015prioritized}. The architecture of the deep neural networks is shown in Table~\ref{tab:montezuma-nn}. The experiment is conducted using Arcade Learning Environment (ALE)~\cite{bellemare2013arcade}. We build customized algorithms based on ALE API to recognize the locations of the agent, the skull, ladders and platforms from pixels and the mapping function $\mathbb{F}$. The intrinsic reward follows (\ref{intrinsic}) with $\phi=1$ and $r=-1$ when the agent loses its life. Extrinsic reward follows (\ref{extrinsic}) where $\psi=100$ and define $r(s,g)=-10$ for $\epsilon>0.9$ to encourage shorter plan. We use hierarchical DQN (hDQN) \cite{kulkarni2016deep} as the baseline.

\noindent{\bf Symbolic Representation.} We formulated domain knowledge in action language ${\cal BC}$ (Fig.~\ref{kr}) based on $6$ pre-defined locations: middle platform ({\tt mp}), right door ({\tt rd}), left of rotating skull ({\tt ls}), lower left ladder ({\tt lll}), lower right ladder ({\tt lrl}), and key ({\tt key}). The input language can be processed by software {\sc cplus2asp} and translated into the input language of \textsc{clingo} for symbolic planning. The function $\mathbb{F}$ maps the symbolic transition to $13$ subtasks (Table~\ref{tab:montezuma-subgoal}). It is worth noting that our subtask definition is
different from hDQN. In hDQN, subtask is associated with an object, but in our work, a subtask is defined as a symbolic transition with initiation set and termination condition mapped from a pair of states whose properties are satisfied by a set of logical literals. Our approach is more descriptive and interpretable, and also makes sub-policy for each subtask to be more easily learned and subtasks more easily sequenced.

\noindent\textbf{Experimental Result.} While data-efficiency is easy to demonstrate quantitatively, interpretability is a qualitative metric. We show that the planning and learning process on subtasks and their sequencing can be easily understood from the figures, providing insights and transparency on how learning optimal behavior progresses under the hood.

We collect data from 10 runs and the shadow in the Fig.~\ref{fig:cumulative},~\ref{fig:success},~\ref{fig:sample} is the variance among these runs. The description of subtasks can refer to both Fig.\ref{fig:final} and Table\ref{tab:montezuma-subgoal}. Only achieving Subtask 3 (picking up the key) and Subtask 7 (opening the right door) can receive the external reward of $+100$ and $+ 300$ respectively, while other subtasks will receive the reward of $0$ from the environment. The learning curve of SDRL (Fig.~\ref{fig:cumulative}) shows that the agent first discovered the plan of collecting key after 0.5M samples by sequencing subtasks 1--3. Intrinsically motivated planning encourages exploring untried subtasks, and by learning more subtasks to move to other locations, the agent finally converges to the maximal cumulative external reward of $400$ around 1.5M samples by sequencing subtasks 1--7 (Fig.~\ref{fig:final}). By comparison, hDQN cannot reliably achieve the score of $400$ around 2.5M samples. The variance of SDRL is smaller than that of hDQN, partially due to the fact that our definition of subtask is easier to learn than the one defined in hDQN, leading to more robust and stable learning.

During the experiment, Subtasks 1--10 are successfully learned by DQNs, with 7 of them being selected in the final solution with achieving success ratio of $100\%$, shown in Fig.~\ref{fig:success}. 
It should be noted that the order of learning subtasks does not depend on the order they appear in the final optimal plan. For instance, Subtask 6 was learned early but appears later in the final optimal plan. This suggests that the subgoals proposed for learning by symbolic planner is activated only when the starting state is satisfied, and once learned, can be easily sequenced and reused in other plans. Subtasks 8--13 are pruned during learning process due to bad extrinsic rewards derived from training performance. Subtask 8, from the lower right ladder to the left of the rotating skull reaches success ratio of $0.9$ but later quickly drops back to $0$, due to the instability of DQN. Subtasks 9 and 10 do not contribute to the optimal plan and are therefore dropped by the planner as well. Subtasks 11 -- 13 are shown to be too difficult to learn in our experiments and discarded by the planner due to poor extrinsic rewards. 
%

\section{Conclusions}\label{sec:conclusion}


In this paper we propose SDRL framework that uses explicitly represented symbolic knowledge to perform high-level symbolic planning based on intrinsic goal and utilizes DRL to learn low-level control policy, leading to improved task-level interpretability for DRL and data-efficiency, which are validated by evaluation on benchmark problems.

\section*{Acknowledgments}
We thank the donation of GPU card from NVIDIA Inc.

\newpage
{\footnotesize
\bibliographystyle{apalike}

}


\section{Appendix}
\label{sec:appendix}
\subsection{Proof of Theorem 1}
\noindent{\textbf{Proof.}} 
When R-learning converges, for any transition $\langle s,a,t\rangle$, the increment terms in (\ref{riter1}) diminish to $0$, which implies 
\begin{align}
R(s,a) &= \max_{a'} R(s,a'), \label{op1:1} \\
\rho^a(s) &= r_e(s,a) - \max_{a'} R(s,a') + \max_{a'} R(t,a') .
\label{op1:2}
\end{align}
Algorithm~\ref{algexec} terminates iff there exists an upper bound of plan quality iff there does not exist a plan with a loop $L$ such that
$
\sum_{\langle s,a,t\rangle\in L} \rho^a(s)> 0,
$
By (\ref{op1:1}) and (\ref{op1:2}), it is equivalent to
$
\sum_{\langle s,a,t\rangle\in L} \left(r_e(s,a)-R(s,a)+R(t,a)\right) \le 0$ iff
$\sum_{\langle s,a,t\rangle\in L} r_e(s,a) - R(s_{|L|},a) + R(s_0,a) \le 0
$
Since $L$ is a loop, $s_{|L|} = s_0$, so 
$
\sum_{\langle s,a,t\rangle\in L} r_e(s,a) \le 0
$
iff any plan $\Pi$ does not have a positive loop of cumulative reward. This is equivalent to the condition that optimal plan exists, which completes the proof.

\subsection{Proof of Theorem 2}
\noindent{\textbf{Proof.}}~By \cite[Theorem 2]{lee13}, $\Pi^*$ is a plan for planning problem $(I, G, D)$. For $\Pi^*$ returned when Algorithm~\ref{algexec} terminates,
$\i{quality}(\Pi) \le \i{quality}(\Pi^*)~\hbox{for~any}~\Pi$ iff $$\sum_{\langle s,a,t\rangle\in\Pi}\rho^a(s) \le \sum_{\langle s,a,t\rangle\in\Pi^*}\rho^a(s).$$ By (\ref{op1:2}), the inequality is equivalent to
$$
\sum_{\langle s,a,t\rangle\in\Pi}r_e(s,a) + R(s_{|\Pi|},a) \le \sum_{\langle s,a,t\rangle\in\Pi^*}r_e(s,a) + R(s_{|\Pi^*|},a).
$$
Since $s_{|\Pi|}$ and $s_{|\Pi^*|}$ are terminal states of each symbolic plan with no options available, we have
$
\sum_{\langle s,a,t\rangle\in\Pi}r_e(s,a)  \le \sum_{\langle s,a,t\rangle\in\Pi^*}r_e(s,a).
$
This completes the proof.

\end{document}